\documentclass[journal]{IEEEtran}

\usepackage[margin=1in]{geometry}
\usepackage{amsmath,amsthm,amssymb,hyperref}
\usepackage{cite}
\usepackage{fancyhdr}
\usepackage{graphicx}
\usepackage{gensymb}
\usepackage{comment}
\usepackage{bm}      
\usepackage{tabularx} 
\usepackage{makecell} 

\usepackage{algorithmic}
\usepackage{textcomp}
\usepackage{hyperref}
\usepackage{wrapfig}
\usepackage{xcolor}
\usepackage{soul}

\usepackage{adjustbox}
\usepackage{lipsum}
\usepackage{caption}
\usepackage{subcaption}
\usepackage{multirow}
\usepackage{array}

\title{Enhancement of Neural Inertial Regression Networks: A Data-Driven Perspective}

\author{
    Victoria Khalfin Fekson, Nitsan Pri-Hadash, Netta Palez, Aviad Etzion, Itzik Klein
    
    The Hatter Department of Marine Technologies, University of Haifa, Israel
}

\date{January 2025}

\begin{document}

\maketitle

\begin{abstract}
Inertial sensors are integral components in numerous applications, powering crucial features in robotics and our daily lives. In recent years, deep learning has significantly advanced inertial sensing performance and robustness.  Deep-learning techniques are used in different domains and platforms to enhance network performance, but no common benchmark is available. The latter is critical for fair comparison and evaluation in a standardized framework as well as development in the field. To fill this gap, we define and thoroughly analyze 13 data-driven techniques for improving neural inertial regression networks. A focus is placed on three aspects of neural networks: network architecture, data augmentation, and data preprocessing. Extensive experiments were made across six diverse datasets that were collected from various platforms including quadrotors, doors, pedestrians, and mobile robots. In total, over 1079 minutes of inertial data sampled between 120-200Hz were analyzed. Our results demonstrate that data augmentation through rotation and noise addition consistently yields the most significant improvements. Moreover, this study outlines benchmarking strategies for enhancing neural inertial regression networks.
\end{abstract}

\IEEEpeerreviewmaketitle

\section{Introduction}
\IEEEPARstart{I}{nertial} sensors play a critical role in modern technology, powering crucial features in our daily lives. From safeguarding our health to enabling the next generation of robotics and autonomous vehicles, these devices are indispensable. Commonly, two types of inertial sensors are considered: accelerometers to measure the specific force vector and gyroscopes to measure the angular velocity vector, enabling monitoring of movement and orientation \cite{groves2015navigation}. An inertial measurement unit (IMU) integrates three orthogonal accelerometers and gyroscopes. Inertial sensors are used in many diverse domains and platforms such as navigation, motion tracking, virtual reality applications, robotics, ground antenna pointing, animal behaviour applications, aerial vehicles, mobile platforms, pedestrian dead reckoning, biomedical and healthcare applications, instrumental buoys for sea monitoring and climate change research, sports, agriculture, internet of things, wearable devices, tracking systems, industrial robotics, and autonomous vehicles. 
In healthcare, IMUs are used to classify diseases by analyzing motion, thanks to the advancement of small and portable units \cite{cuesta2010use, engelsman2022measurement}. In navigation, IMUs are integral to diverse applications in aerial vehicles, autonomous vehicles, and pedestrian movement \cite{cohen2024inertial, klein2022data}, due to their self-contained nature and capability to provide crucial navigation data without external references \cite{groves2015navigation}. 
In sports, accelerometers and gyroscopes measure performance during training or competition. Their small size allows them to integrate into an athlete's outfit during physical activity easily \cite{camomilla2018trends, worsey2019inertial}. 
While IMUs find widespread application across diverse fields, their inherent limitations pose considerable challenges to achieving high levels of accuracy and reliability. These limitations primarily stem from factors such as sensor noise, intrinsic biases, temporal drift, and susceptibility to environmental conditions. The cumulative impact of these inaccuracies can lead to substantial errors that propagate and magnify over time, potentially compromising the validity of measurements and subsequent analyses\cite{groves2015navigation, shaeffer2013mems}. 

\noindent Deep learning (DL) has achieved breakthroughs across domains like computer vision, NLP, and time-series analysis, driven by its ability to process large datasets and extract features automatically, surpassing model-based and classical machine learning methods \cite{sarker2021deep, malhotra2023recent}. In recent years, DL has emerged as a transformative approach to addressing challenges in inertial sensing, demonstrating significant advancements in areas such as sensor calibration, denoising, and navigation across diverse platforms \cite {cohen2024inertial, klein2022data, chen2024deep}. DL models, such as convolutional neural networks (CNNs), recurrent neural networks (RNNs) along with their variants (long short-term memory (LSTM) and gated recurrent unit (GRU) networks), as well as attention-based architectures, have demonstrated exceptional effectiveness in capturing complex non linear patterns from IMU data, frequently outperforming all other methods. For instance, DL-based calibration techniques have been employed to mitigate deterministic and random errors in low-cost IMUs, with CNN architectures successfully reducing bias and noise in accelerometer and gyroscope readings \cite{li2022calib}. Similarly, learning-based approaches, such as RNN, LSTM, and GRU, for denoising stationary accelerometer signals, outperform traditional signal processing techniques \cite{engelsman2023data}. 
Pedestrian dead-reckoning (PDR) applications have also benefited from DL, with studies focused on learning and reconstructing pedestrian trajectories from raw IMU data collected by smartphones \cite {asraf2021pdrnet, chen2020deep}. In underwater navigation, DL methods involve the fusion of data from Doppler velocity logs (DVLs) and inertial data to improve accuracy \cite{zhang2020navnet, or2023pronet}. 
Aerial navigation has also seen notable improvements through DL, particularly in visual-inertial odometry where architectures like VIO-DualProNet combine CNN blocks to process IMU data for covariance noise estimation integration with visual data \cite {solodar2024vio}. Also, approaches like AbolDeepIO utilize LSTM networks for 3D position and orientation estimation \cite{esfahani2019aboldeepio}. These examples underscore the versatility and efficacy of DL in improving the robustness and adaptability of inertial navigation systems across varied applications and environments.

\noindent However, despite these advancements, several challenges remain. One of the key obstacles lies in the variability of sensor noise, body placement of devices, and environmental conditions, which can introduce inaccuracies in the inertial measurements \cite{cohen2024inertial, groves2015navigation}. Moreover, the computational cost and data-hungry nature of deep networks often require innovative training methodologies, data augmentation strategies, and optimization techniques to generalize effectively across various conditions and users \cite{bengio2013deep, kawaguchi2017generalization}. Finally, deep-learning techniques are used in different domains and platforms to enhance network performance, but no common benchmark is available. The latter is critical for fair comparison and evaluation in a standardized framework as well as development in the field.

\noindent To fill this gap, we define and thoroughly analyze 13 data-driven techniques for improving neural inertial regression networks. To this end, we distinguish between three major approaches commonly used in data-driven theory: network architectural design, data augmentation, and data preprocessing. In network architectural design, we examine multi-head architectures and various loss functions. For data augmentation, we investigate rotation, additive bias, and additive noise techniques. Our data preprocessing methods encompass inertial noise handling through both denoising and noise addition, data normalization, and detrending.
We evaluate these techniques across six real-world recorded datasets (with eight sub-datasets) collected from diverse platforms including quadrotors, doors, pedestrians, and mobile robots. In total, we analyze over 1079 minutes of recorded inertial data sampled between 120-200Hz, all used in supervised learning regression problems. Unlike many studies that examine isolated techniques in a specific application, this research provides an extensive evaluation of multiple strategies across various applications and platforms, offering a broader perspective on which methods can consistently improve model accuracy and which may be more effective in specific scenarios. By applying these techniques in varied real-world contexts, this paper offers practical insights for implementing neural networks in inertial sensing applications.\\
Among all the techniques evaluated, data augmentation through rotation and noise addition emerged as the most consistently effective methods to improve the performance of neural inertial networks. Moreover, this study outlines benchmarking strategies for enhancing neural inertial regression networks.

\noindent The rest of the paper is organised as follows: Section \ref{sec:methodology} gives our methodology including the baseline network with the 13 data-driven techniques used in this paper. Section \ref{sec:res} presents our results and Section \ref{sec:conclusions} gives the conclusions of this study. 

\section {Methodology} \label{sec:methodology}
\noindent This section details the methodological framework employed in this study. It starts by presenting the baseline network architecture used to evaluate data-driven paradigms. Next, it discusses the paradigms themselves, including architecture design, data augmentation, and data preprocessing techniques. Finally, it provides an overview of the datasets used, emphasizing their significance and outlining the preprocessing steps performed to prepare them for model training and evaluation.

\subsection {Baseline network} \label{sec:networks}
\begin{figure*}[!htb]
    \centering
    \includegraphics[width=\textwidth, height=2cm]{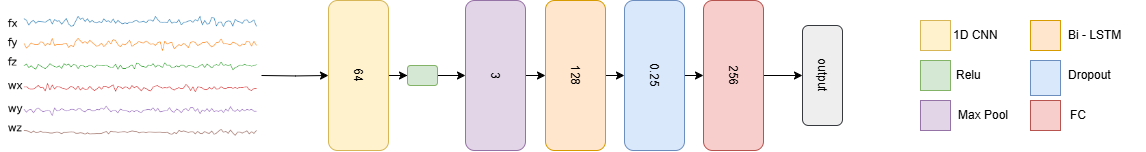}
    \caption{Baseline network architecture. The output differs between each task and dataset.}
    \label{fig:network}
\end{figure*}

\noindent We used a network architecture inspired by the model presented in \cite{silva2019end}. This architecture integrates convolutional neural network (CNN), bidirectional long short-term memory (Bi-LSTM) layer, and fully connected (FC) layer, as depicted in Figure \ref{fig:network}.
The network input is a time-series signal of inertial readings:
\begin{equation}
x_t = [f_x, f_y, f_z,\omega_x,\omega_y,\omega_z]^{T}\in\mathbb{R}^6
\end {equation}
where \(f\) is the specific force vector as measured by the accelerometers, \(\omega\) is the angular velocity vector measured by the gyroscopes, and \(t\) is the time index. Note that the measurements are expressed in the sensor coordinate frame, and we omit it from the mathematical notation for brevity.

\noindent The input signal is passed through one layer of 1D convolutional layer with \(F=64\) filters, a kernel size \(k = 5\), and stride \(s=1\). The convolution operation for each feature map \(y_f(t)\) is given by:
\begin{equation}
\hspace{-0.2cm} 
y_f(t) = \mathrm{ReLU}\left(\sum_{c=1}^6 \sum_{i=1}^k w_f^{(c,i)} x_c(t+i-1) + b_f\right)
\end {equation}
where \(w_f^{(c,i)}\) are the learnable weights for filter \(f\), channel \(c\), and offset \(i\), \(b_f\) is the bias term, and \(x_c(t+i-1)\) is the input for channel \(c\) at time step \(t\).
The ReLU is a nonlinear activation function defined by:
\begin{equation}
\mathrm{ReLU}(x) = \max(0, x)
\end {equation}
The output from the convolution layer is then passed into a max pooling layer to reduce its temporal dimension:
\begin{equation}
y_f^{\text{p}}(t) = \max_{i \in (0, d)} y_f(t+i),
\end {equation}
where \(y_f^{\text{p}}(t)\) represents the output of the max pooling layer at time step \(t\), \(y_f(t+i)\) is the value of the feature map \(f\) at time \(t+i\) before pooling, and \(d = 3\) is the pooling depth, which defines the size of the window used for pooling. 

\noindent The output of the pooling layer is fed into a Bi-LSTM layer. The Bi-LSTM processes the sequence in both forward and backward directions, producing hidden states for each time step:
\begin{equation}
\overrightarrow{\bm{h_t}} = \text{LSTM}({\bm{x_t}}, \overrightarrow{\bm{h_{t-1}}}), \quad
\overleftarrow{\bm{h_t}} = \text{LSTM}({\bm{x_t}, \overleftarrow{{\bm{h_{t+1}}}}})
\end {equation}
where \( \overrightarrow{\bm{h_t}} \) is the hidden state at time step \( t \) computed by the forward LSTM layer based on the current input \( x_t \) and the previous forward hidden state \( \overrightarrow{\bm{h_{t-1}}} \), and \( \overleftarrow{\bm{h_t}} \) is the hidden state at time step \( t \) computed by the backward LSTM layer using the current input \( x_t \) and the next backward hidden state \( \overleftarrow{\bm{h_{t+1}}} \).
The concatenated hidden states are then passed through a dropout layer to prevent overfitting during training. 
\begin{equation}
\bm{h_t^{\text{dropout}}} = \bm{h_t} \odot \bm{r}, \quad \bm{r} \sim \text{Bernoulli}(1-p)
\end {equation}
where \( \bm{h_t^{\text{dropout}}} \) is the output after applying dropout, \( \bm{r} \) is a binary mask sampled from a Bernoulli distribution with probability \( 1-p \), where \( p = 0.25\) is the dropout rate, and \(\odot\) denotes element-wise multiplication.
The output from the dropout layer is then passed through a FC with 256 neurons:
\begin{equation}
\bm{y_{\text{FC}}} = \bm{h_t^{\text{dropout}}}\mathbf{W}_{\text{FC}} + \bm{b}_{\text{FC}} 
\end{equation}
where \( \mathbf{W}_{\text{FC}} \) is the weight matrix of the fully connected layer and \( \bm{b}_{\text{FC}} \) is the bias vector.

\noindent This compact network is specifically designed to highlight the improvements achieved through the experimental paradigms, rather than through the network architecture itself. While this helps isolate the effects of the paradigms, it also means the architecture may not fully leverage advanced design strategies that could further enhance performance.

\noindent The Adam optimizer \cite{kingma2014adam} was used with a learning rate of 0.001. The training was done on a single NVIDIA GeForce RTX 4090 GPU with a batch size of 64 samples. The number of epochs varied for each dataset, depending on factors such as convergence and running time. This will be discussed in Section \ref{sec:datasets} for each dataset. Additionally, we explored various types of loss functions, which will be covered in Section \ref{sec:loss_functions}.

\subsection{Data-driven perspectives on inertial data} \label{sec:paradigms}
\begin{figure*}[!htb]
    \centering
    \includegraphics[width=\textwidth, height=2.5cm]{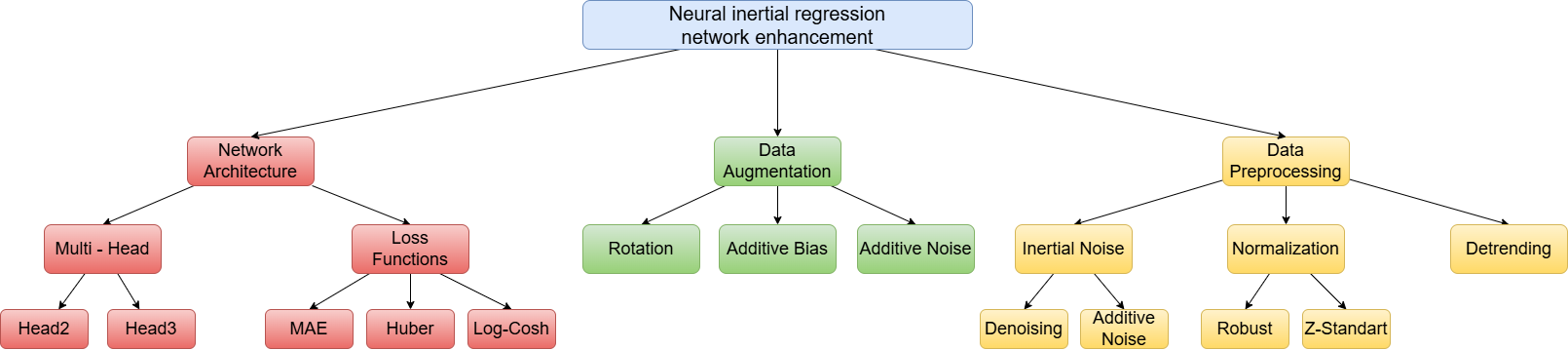}
    \caption{Taxonomy tree of data-driven techniques used for enhancing neural inertial networks.}
    \label{fig:tree}
\end{figure*}

\noindent This section discusses the approaches used in the current study to process and optimize inertial neural networks. Figure \ref{fig:tree} gives the hierarchical structure of the methods addressed in this paper. The tree diagram illustrates how different approaches are organized and related, providing an overview of the various data-driven techniques.

\subsubsection{Network Architectural Design}
\begin{figure*}[!htb]
    \centering
    \includegraphics[width=\textwidth]{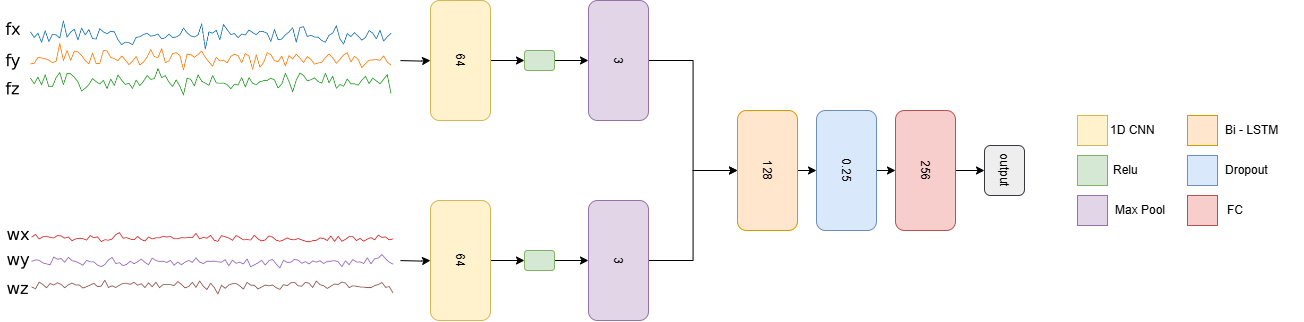}
    \caption{Head2 architecture. One head receives accelerometer data and the other receives gyroscope data.}
    \label{fig:two_head}
\end{figure*}

\begin{figure*}[!htb]
    \centering
    \includegraphics[width=\textwidth, height=4.5cm]{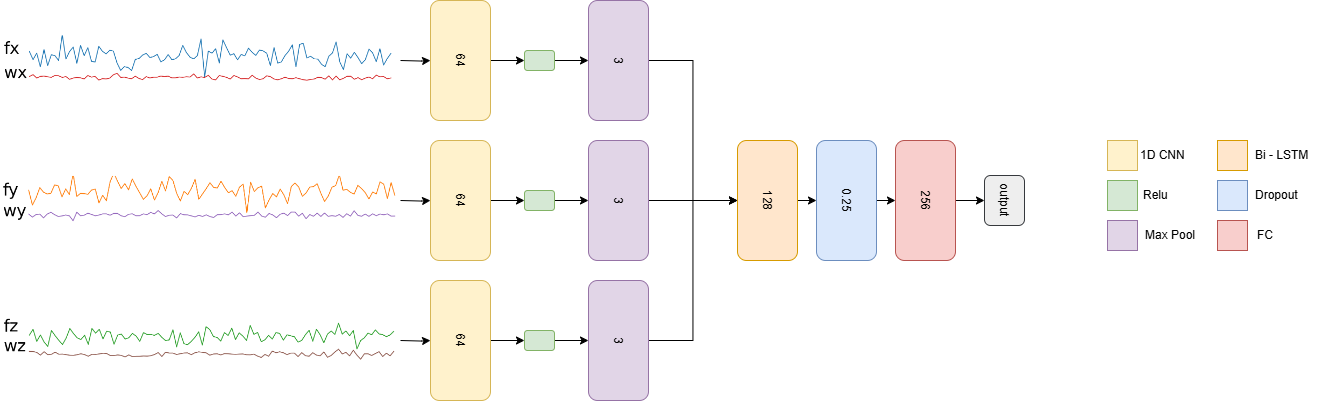}
    \caption{Head3 architecture. Each head receives the accelerometer and gyroscope readings along the x,y, and z axes.}
    \label{fig:three_head}
\end{figure*}

\paragraph{\textbf{Multi-Head Network}} \label{sec:multi_head}
Various deep learning architectures have been applied to inertial data processing tasks. Among these architectures, some utilize a single-head approach where both accelerometer and gyroscope data are passed through a single processing unit \cite{herath2020ronin,esfahani2019aboldeepio, shavit2021boosting}. Alternatively, a two-head architecture has been employed, where accelerometer data is processed separately from gyroscope data \cite{silva2019end,liu2023smartphone}. There is still uncertainty in the literature regarding which architecture demonstrates superior performance.
Therefore, we offer two multi-head architectures for the evaluation:
\begin{itemize}
\item \textbf{Head2} 
Here we use two heads, one for the accelerometer readings and one for the gyroscopes. as shown in Figure \ref{fig:two_head}. 
\item \textbf{Head3} 
Here we use three heads, one for each inertial axis. That is, we couple the accelerometer and gyro x-axis in a single head and do the same for y and z axes. as presented in Figure \ref{fig:three_head}.
\end{itemize}

\paragraph{\textbf {Loss Functions}} \label{sec:loss_functions}
In the context of loss functions, the literature contains many different loss functions \cite{wang2020comprehensive}, both in general contexts and in the analysis of inertial data. 
We examined four types of loss functions to observe their impact on the inertial data.

\noindent To that end, we adopt the following notation:
$y$ is the true value, $\hat{y}$ is the predicted value, $n$ is the number of samples in a batch, and $L$ is the loss function.
The overall loss is computed for all the batches in the training set.\\

\begin{itemize}
\item \textbf{Mean Squared Error (MSE) \cite{wang2020comprehensive}:} computes the average of the squared differences between actual values and predicted values:
\begin{equation}
    L_{MSE} = \frac{1}{n} \sum_{i=1}^{n} (y_i - \hat{y_i})^2 \label{eq:MSE_eq}
\end{equation}

\item \textbf{Mean Absolute Error (MAE) \cite{wang2020comprehensive}:} computes the average of the sum of absolute differences between actual values and predicted values: 
\begin{equation}
    L_{MAE} = \frac{1}{n} \sum_{i=1}^{n} |y_i - \hat{y_i}| \label{eq:MAE_eq}
\end{equation}

\item \textbf{Huber \cite{wang2020comprehensive}:} is a combination of the MSE and MAE loss functions. The Huber loss behaves like MSE when the error is small and like MAE for large errors, providing robustness against outliers. The decision is controlled by a threshold hyperparameter, $\delta$, that one can tune. 
The Huber loss function is defined as: 

\begin{equation}
\hspace{-0.5cm}
\begin{adjustbox}{width=0.85\columnwidth}
    $L_{Hu} =  
    \begin{cases}
        \frac{1}{n} \sum_{i=1}^{n} \frac{1}{2}(y_i - \hat{y}_i)^2, & \text{if } |y_i -  \hat{y}_i| \leq \delta \\
        \frac{1}{n} \sum_{i=1}^{n} (\delta |y_i - \hat{y}_i| - \frac{1}{2} \delta^2), & \text{otherwise}
    \end{cases}$
\end{adjustbox}
\end{equation}

\item \textbf{Log-Cosh \cite{wang2020comprehensive}:} defined as the logarithm of the hyperbolic cosine of the prediction error: 
\begin{equation}
    L_{log-cosh} =  \frac{1}{n} \sum_{i=1}^{n} log(cosh(\hat{y_i}-y_i)) \label{eq:Log_Cos_eq}
\end{equation}
Similar to the Huber loss, it has the advantage of being less sensitive to outliers.

\end{itemize}

\subsubsection{Data Augmentation} \label{subsec:aug}
To train deep learning networks effectively, a substantial amount of data is necessary to achieve good performance \cite {he2016deep}. However, collecting extensive datasets is often challenging.  
To address this, data augmentation techniques have been developed across various research fields to generate additional training data. 
For inertial data, some of the augmentation techniques include rotation, permutation, scaling, cropping, additive bias, and additive noise \cite{um2017data}. 
In this paper, we utilize three augmentation techniques: data rotation, additive bias, and additive noise. We examine each of the three separately by adding the new data to the existing training set.
\paragraph{\textbf{Rotation}}
Rotation augmentation enhances time series datasets by applying rotational transformations to the input data. This approach maintains the relationships between features while introducing variability and improving model generalization. 
The rotation transformation is defined by:
\begin{align}
    \tilde{\bm{x}} &= \mathbf{R} \cdot \bm{x} \label{eq:rot} 
\end{align}
where $\bm{x}$ is a sample of the inertial measurements vector, $\mathbf{R}$ is the rotation matrix, and $\tilde{\bm{x}}$ is the rotated vector.
\noindent As we employ different rotation matrices for each dataset, we give the specific values in the next section.

\paragraph{\textbf{Additive Bias}}
Introducing bias is a straightforward and effective technique for augmenting inertial time series data by adding a constant offset to the input. This simulates real-world conditions, such as sensor calibration errors or baseline shifts. The bias is randomly sampled from a Gaussian distribution with a zero mean and a dataset-specific standard deviation for each axis. The augmentation process is mathematically described by:
\begin{align}
    \tilde{\bm{x}} &= \bm{x} + \bm{b} \label{eq:add_bias}
\end{align}
where $\bm{x}$ is a sample of the inertial measurements vector, $\bm{b}$ is a random bias vector, and $\tilde{\bm{x}}$ is the vector after the manipulation.

\noindent As we employ different bias values for each dataset, we give the specific values in the next section.

\paragraph{\textbf{Additive  noise}}
Noise addition is a widely used data augmentation technique that enhances the robustness of machine learning models by introducing random perturbations into the data. For inertial time series, noise simulates real-world factors such as sensor inaccuracies, environmental disturbances, or random fluctuations. The noise is drawn from a Gaussian distribution with a mean of zero and a dataset-specific standard deviation. Mathematically, the noise addition process can be expressed as:
\begin{align}
    \tilde{\bm{x}} &= \bm{x} + \mathcal{N}(0, \sigma^2) \label{eq:aug_add_noise}
\end{align}
where \( \bm{x} \) represents the original input sample, \( \mathcal{N}(0, \sigma^2) \) denotes the Gaussian noise with zero mean and variance \( \sigma^2 \), and \( \tilde{\bm{x}} \) is the augmented sample with the added noise.

\noindent As we employ different noise values for each dataset, we give the specific values in the next section.

\subsubsection{Data Preprocessing Techniques} 
\paragraph{\textbf{Inertial Noise}}
Two approaches for handling inertial noise are introduced: denoising and noise addition.
\begin{itemize}
\item \textbf{Denoising} is commonly explored to assess its potential impact on improving the quality of measurements and reducing the influence of sensor noise \cite {khaddour2021survey}. Conventional signal processing approaches, as well as recent learning-based techniques, have been investigated to address this challenge.

\noindent One of the traditional signal processing methods, discussed extensively in the literature, is the moving average (MA) technique \cite{engelsman2023data}. MA techniques serve as efficient smoothing filters, leveraging errors (residuals) from previous forecasts \cite{gonzalez2018statistical}. Research on denoising techniques for inertial sensors, which started in the late 90s, has looked into how MA filters can help clean up the data before using it for things like figuring out Euler angles or navigation states. However, determining the optimal window size for MA filters is often heuristic and relies on the characteristics of the dataset being processed.

\noindent The denoising process using the moving average filter can be expressed by the following equation:
\begin{equation}
\tilde{x}_t = \frac{1}{n}\sum^{n-1}_{i=0} x_{t+i} \label{eq:ma_eq}
\end{equation}
where \( n \) is the window size, \( x_{t+i} \) is the accelerometer or gyroscope measurements at each time step within the window, and \(\tilde{x}_t\) is the denoised value at time \(t\).

\noindent In our experiments, we utilized the MA technique to filter noise from the IMU data. The window size for the moving average filter was selected empirically to optimize its performance on the dataset.

\item \textbf{Noise addition} addresses the vulnerability of IMUs to both random and deterministic noise, which can compromise the accuracy and reliability of the collected data \cite{klein2022data}. By introducing controlled random noise, robustness can be enhanced, and outliers effectively mitigated. Abolfazli Esfahani et al. proposed the addition of random Gaussian noise to data before processing it with a deep neural network, demonstrating improved model accuracy across two experimental setups \cite{esfahani2019aboldeepio, esfahani2019orinet}.

\noindent In our study, we adopted the Gaussian noise addition technique to introduce controlled noise into the IMU data. By modulating the mean and standard deviation parameters of the Gaussian distribution, we simulated realistic noise conditions to assess the system's performance under varying noise levels. 
The following equation describes the operation:  
\begin{align}
    \tilde{\bm{x}} &= \bm{x} + \mathcal{N}(0, \sigma^2) \label{eq:pre_add_noise}
\end{align}
where \( \bm{x} \) represents the original input sample, \( \mathcal{N}(0, \sigma^2) \) denotes the Gaussian noise with zero mean and variance \( \sigma^2 \), and \( \tilde{\bm{x}} \) is the augmented sample with the added noise.
\end{itemize}
\paragraph{\textbf{Data Normalization}}
Normalization is a data preprocessing technique used to scale the dataset features to a common range without distorting differences in the ranges of values. The primary goal of normalization is to ensure that each feature contributes equally to the analysis, especially in machine learning models where the magnitude of data values can significantly impact the performance and convergence of algorithms \cite {cabello2023impact}. 
In our analysis, we examined two commonly used normalization techniques \cite {lima2023large}:  

\vspace{3mm}

\begin{itemize}
\item \textbf{Z-standard normalization} – commonly referred to as standardization, involves adjusting each value in the dataset so that the mean becomes 0 and the standard deviation becomes 1, such that:
\begin{equation}
    x_{z-score} = \frac{x - \mu}{\sigma} \label{eq:z_standard_eq}
\end{equation}

where \( x_{z-score}\) is a single sample, \( \mu \) is the mean of the data, and \( \sigma \) is its standard deviation.

\item \textbf{Robust normalization} – Outliers can disproportionately impact the scaling of a feature, leading to distortion in standard normalization. Robust scaling is more effective for features containing outliers. The operation is defined by the following equation:
\begin{equation}
    x_{\text{Robust}} = \frac{x - x_{\text{median}}}{x_{\text{IQR}}} \label{eq:robust_eq}
\end{equation}
where \( x \) is a single sample, \( x_{\text{median}} \) is the median of the data, and \( x_{\text{IQR}} \) is the interquartile range, which measures the spread of the data.
\end{itemize}

\paragraph{\textbf{Detrending}}
Detrending is a signal processing method employed to eliminate trends from data, allowing for a better focus on the underlying dynamics of the signal. Previous research has shown that detrending inertial data can lead to improved results \cite{or2023pronet}. The most common approach to detrending is linear detrending, which involves fitting a straight line to the data, described by:
\begin{equation}
    \hat{\bm x}(t) = a \cdot \bm x(t) + b \label{eq:linear_detrend_eq}
\end{equation}
where \( \hat{\bm x}(t) \) is the trend of time point \( t \), \( \bm x(t) \) is the original sample at time point \( t \), \( a \) is the slope, and \( b \) is the intercept of the line. \\
The detrended signal is obtained by subtracting this trend from the original signal: 
\begin{equation}
    \bm x_{\text{detrend}}(t) = \bm x(t) - \hat{\bm x}(t). \label{eq:detrend_eq}
\end{equation}
where \( \bm x(t) \) is the original signal at time point \( t \), and \( \hat{\bm x}(t) \) is the fitted trend.

\subsection {Datasets} \label{sec:datasets}
\noindent We used six datasets (eight sub datasets) collected from a variety of fields with different platforms such as quadrotors, doors, pedestrians, and mobile robots. In total, we used 1079 minutes of recorded inertial data sampled between 120-200Hz.
All datasets were used for regression learning problems.

\subsubsection{QuadNet - Quadrotor Data} \label{quadnet_data}
The quadrotor dataset is fully described in \cite{hurwitz2023quadrotor, hurwitz2024deep} and publicly available under the ANSFL GitHub at:
\url {https://github.com/ansfl/Quadrotor-Dead-Reckoning-with-Multiple-Inertial-Sensors}.
The data was collected using a DJI Phantom 4 quadrotor as the main platform. This quadrotor was equipped with onboard D-RTK to accurately track its movements. Additionally, it had four Xsens DOT IMUs onboard. Data collection was performed at 120 Hz. For our study, we focused solely on data from IMU1 in both vertical and horizontal datasets. The purpose of collecting this dataset was to mimic pedestrian movement, enabling the exploration of PDR techniques. Obtaining that requires flying the quadrotor in a periodic trajectory. In this study, we utilized this dataset to predict the quadrotor's distance in the x,y plane. A window size of 120 and a stride of 60 were applied in the data preprocessing stage.

\subsubsection{EuRoC\_MAV - Quadrotor Data } 
\label{euroc_mav}
The EuRoC MAV dataset is fully described in \cite{burri2016euroc} and publicly accessible through the autonomous systems lab (ASL) datasets site at: \url {https://projects.asl.ethz.ch/datasets/doku.php?id=kmavvisualinertialdatasets#the_euroc_mav_dataset}
This dataset provides angular velocity and raw acceleration data recorded with an AscTec Firefly hex-rotor helicopter\cite{achtelik2012design}, equipped with an ADIS16448 IMU at a sampling rate of 200 Hz. Precise and synchronized ground truth of 3D position and orientation are also provided. The dataset contains 11 sequences, a recording time of approximately 23 min and a total trajectory length of 894 m. In this study, we chose nine trajectories for training and two for testing. A window size of 200 and a stride of 50 were applied in the data preprocessing stage.

\subsubsection{DoorINet - Doors Data} \label{doorinet_data}
The DoorINet dataset is fully described in \cite{zakharchenko2024doorinet} and is publicly accessible via the ANSFL GitHub repository at: \url{https://github.com/ansfl/DoorINet}{https://github.com/ansfl/DoorINet}. 
The dataset was collected using two distinct types of IMUs: the Memsense MS-IMU3025 and the Movella Xsens DOT. The Memsense MS-IMU3025 was utilized for generating ground-truth (GT) readings, operating at a recording frequency of 250Hz, while the Movella Xsens DOT IMUs, recorded at 120Hz, served as the units under test. All IMUs were positioned on a door, and the dataset was curated to predict the door's heading angle. a window size of 20 and a stride of 20 were applied in the data preprocessing stage. 

\subsubsection{RIDI - Pedestrian Data } \label{ridi}
The RIDI dataset is fully described in \cite{yan2018ridi} and is publicly accessible through Kaggle at: \url {https://www.kaggle.com/datasets/kmader/ridi-robust-imu-double-integration}. 
The dataset comprises IMU sensor measurements and motion trajectories from human subjects and four common device positions: in a leg pocket, in a bag, held by a hand, or on the body. 
Over 150 minutes of data were collected at a sampling rate of 200 Hz from ten individuals across the previously mentioned four smartphone placements, encompassing various motion types such as walking forward/backwards, side motion, or acceleration/deceleration. The recordings were collected with a Google Tango phone, Lenovo Phab2 Pro.
In this study, we utilized the RIDI dataset to predict the pedestrian's location in x and y coordinates.
A window size of 200 and a stride of 10 were applied in the data preprocessing stage. 

\subsubsection{RoNIN - Pedestrian Data } 
\label{ronin}
The RoNIN dataset is fully described in \cite{herath2020ronin} and is publicly accessible through the official research website at: \url {https://ronin.cs.sfu.ca/}.
The dataset is a comprehensive pedestrian movement dataset consisting of 42.7 hours of data collected from 100 subjects using three different Android devices, which the subjects held naturally. The ground truth was obtained using a 3D tracking phone (Asus Zenfone AR) attached to the body via a harness. Data collection was performed at a frequency of 200 Hz. The dataset is divided into two groups: Group 1 consists of 85 subjects, and Group 2 contains the remaining 15 subjects. Group 1 is further split into training, validation, and testing subsets, whereas Group 2 is utilized to assess the model's generalization capability to new, unseen subjects.
In this study, we chose a subset of 45 subjects from the dataset to conduct our experiments and shorten the running time. Specifically, we used 20 subjects from Group 1 for training the model, 5 subjects from Group 1 for validation, and 10 subjects from Group 1 for testing with data the model has seen before. Additionally, we selected 10 subjects from Group 2 to test the model's performance on new, unseen data. 
A window size of 200 and a stride of 100 were applied in the data preprocessing stage.

\subsubsection{MoRPI - Mobile Robots } 
\label{morpi}
The MoRPI dataset is fully described in \cite{etzion2024snake} publicly available under the ANSFL GitHub at:
\url {https://github.com/ansfl/MoRPINet}.
The dataset consists of data from a mobile robot, specifically the STORM Electric 4WD Climbing Car. It includes trajectories from a Javad SIGMA-3N RTK sensor operating at 10Hz, as well as recordings from five different Movella DOT IMUs at 120Hz. A total of sixteen distinct trajectories were captured during field experiments. In this study, we selected two IMUs (IMU1 and IMU2) from each trajectory. The data was used to predict distance in the x,y plane, with a window size of 360 and a stride of 60 applied during the preprocessing stage.

\subsubsection{Summary} 
\label{datasets_summary}
Table \ref{tbl:datasets_info} presents a summary of the dataset characteristics and training parameters. Altogether, we analyze six datasets (including eight sub-datasets), comprising a total of 1079 minutes of recorded inertial data sampled at frequencies ranging from 120 to 200 Hz.

\begin{table*}[!thb]
    \centering
    \caption{Dataset information and parameters.}
    \label{tbl:datasets_info}
    \begin{tabular}{|l|l|l|l|l|l|l|}
    \hline
    \makecell[l]{\textbf{Dataset}} &  \makecell[l]{\textbf{Platform}} & \makecell[l]{\textbf{Sampling} \\ \textbf{Rate (Hz)}} & \makecell[l]{\textbf{Train} \\ \textbf{Length (min)}} & \makecell[l]{\textbf{Test} \\ \textbf{Length (min)}} & \makecell[l]{\textbf{Window Size/} \\ \textbf{Stride}} & \makecell[l]{\textbf{Epochs}} \\ \hline
    \makecell[l]{QuadNet \\ Horizontal} & Quadrotor & 120 & 13.6  & 1.34 & 120/60 & 150 \\ \hline
    \makecell[l]{QuadNet \\ Vertical} & Quadrotor & 120 & 10.05  & 1.29 & 120/60 & 150 \\ \hline
    EuRoC\_MAV    & Quadrotor    & 200 & 18.2 & 4.24 & 200/100 & 150 \\ \hline
    DoorINet       & Doors  & 120 & 119.12 & 271.83 & 20/20 & 100 \\ \hline
    RIDI          & Pedestrian    & 200 & 77.24 & 36.64 & 200/100 & 100 \\ \hline
    RoNIN & Pedestrian & 200 & 205.58 & \makecell[l]{Seen-103.17 \\ Unseen-93.09} & 200/10 & 120 \\ \hline
    MoRPI & Mobile Robot & 120 & 105.52 & 18.14 & 360/60 & 80 \\ \hline
    \textbf{Total Length} & & & \textbf{549.31} & \textbf{529.74} & & \\ \hline
    \end{tabular}
\end{table*}

\section{Results} \label{sec:res}
\noindent In the upcoming section, we outline the outcomes of our diverse experiments.
To address the inherent randomness in training deep learning networks \cite{zhuang2022randomness} and enhance the reliability of the results, experiments for various tasks were conducted multiple times with different random seeds. Each experiment was executed 30 times, except for the DoorINet experiments, which were conducted 20 times due to performance constraints.

\subsection{Evaluation Metric}

\noindent We utilized the Root Mean Square Error (RMSE) as the evaluation metric to assess the performance of the models. The RMSE is calculated using the following equation:

\begin{equation}
\text{RMSE} = \sqrt{\frac{1}{n} \sum_{i=1}^{n} (y_i - \hat{y}_i)^2}
\end{equation}

\noindent where \(y_i\) represents the actual values, \(\hat{y}_i\) denotes the predicted values, and \(n\) is the total number of samples. The improvements are presented as a percentage, calculated based on the reduction or increase in RMSE relative to the baseline.

\subsection{Network Architectural Design}
\subsubsection{Multi-Head Network} \label{res:multi_head}
\begin{figure}[!htb]
    \centering
    \includegraphics[width=\linewidth]{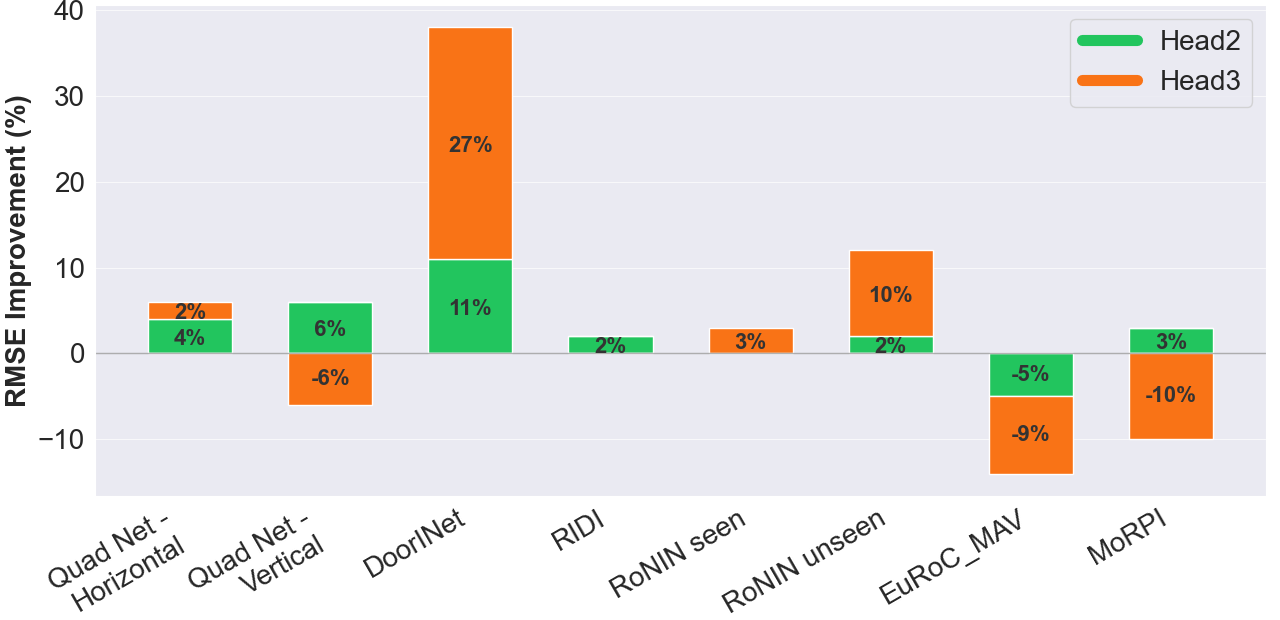}
    \caption{\small RMSE improvement in percentages for each dataset using the multi-head architecture.}
    \label{fig:multi_head_res_sum}
\end{figure}

The effect of different multi-head architectures is evaluated across all datasets. A summary of the results in terms of RMSE improvement is given in Figure \ref{fig:multi_head_res_sum}. 
The Head2 architecture demonstrated improvements in seven out of eight datasets, achieving an average RMSE improvement of 3\%, while the Head3 architecture showed improvements in five out of eight datasets, with an average gain of 2\%. Both multi-head architectures performed exceptionally well on the DoorINet dataset, achieving improvements of 11\% and 27\% for Head2 and Head3, respectively. However, they negatively impacted the EuRoC dataset, with performance declines of -5\% and -9\%, respectively.
In comparing the two multi-head architectures, the Head3 architecture shows significant advantages over Head2 in specific tasks, such as a 10\% improvement on the RoNIN unseen dataset. However, it also displays greater variability across datasets, with performance drops of -10\% on the MoRPI dataset and -6\% on the QuadNet vertical. Conversely, the Head2 architecture offers more consistent performance gains across most datasets compared to the baseline.

\subsubsection{Loss Functions} \label {res:loss_functions}
\begin{figure}[!htb]
    \centering
    \includegraphics[width=\linewidth]{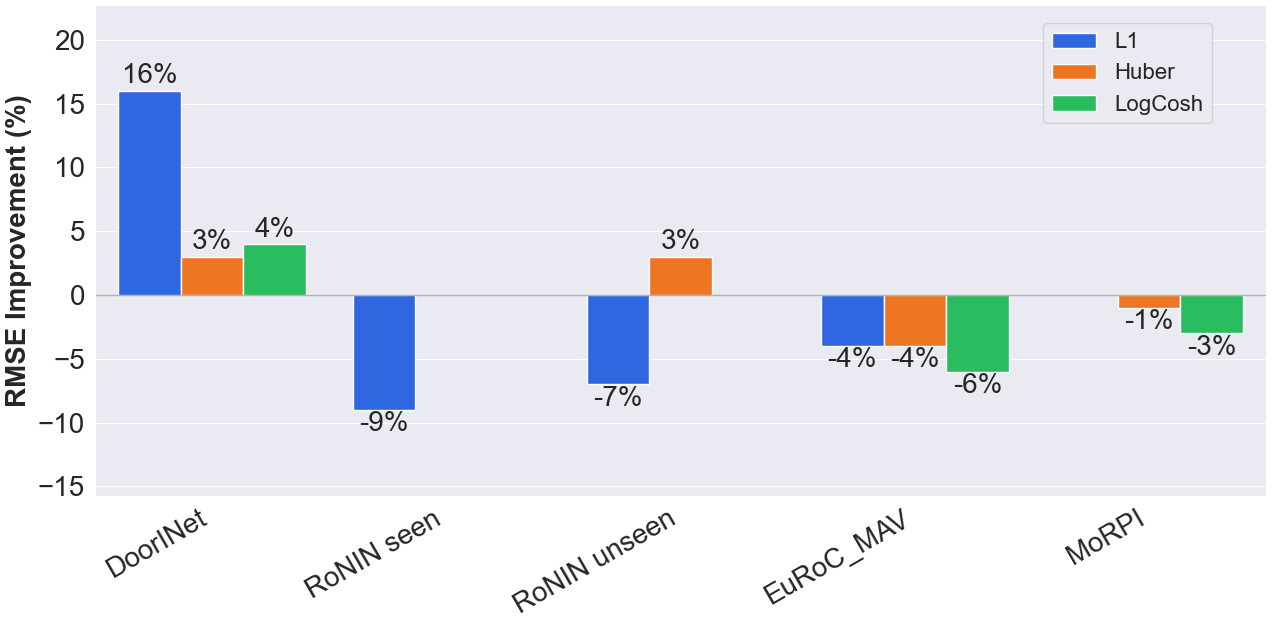}
    \caption{\small RMSE improvement in percentages for each dataset when using various loss functions compared to MSE loss.}
    \label{fig:loss_res}
\end{figure}

The results for the three different loss functions are summarized in Figure \ref{fig:loss_res}. There, each loss function is compared to the MSE loss. For the Huber loss function, a value of $\delta = 1.0$ was used. Notice that the figure shows the results only for DoorINet, RoNIN, EuRoC\_MAV, and MoRPI datasets, as in all others the improvement or degradation of the results is less than 3\%. 
The L1 loss demonstrated improvements in one out of eight datasets, with an average change of -1\%, the Huber loss showed gains in three out of eight datasets, averaging 0\%, and the LogCosh loss improved performance in one out of eight datasets, with an average change of -1\%.
On average, no consistent improvement was observed over the MSE loss. However, in specific cases, certain loss functions demonstrated notable enhancements. For instance, on the DoorINet dataset, the L1 loss achieved a significant 16\% improvement, while the Huber loss improved performance by 3\% on the RoNIN unseen dataset.

\subsection{Data Augmentation} \label {res:data_aug}
\noindent Three types of training sets were constructed:
1) \textbf{Bias}: Consists of the original data with additional data created by adding a bias to the original data.
2) \textbf{Noise}: Consists of the original data with additional data created by adding noise to the original data.
3) \textbf{Rotation}: Consists of the original data with additional data created by applying rotations to the original data.\\
The rotated data was created using three rotation matrices, each emphasising a different axis: 
\begin{equation}
T_1 = \begin{bmatrix}
\cos\left(\frac{\pi}{6}\right) & \sin\left(\frac{\pi}{6}\right) & 0 \\
-\sin\left(\frac{\pi}{6}\right) & \cos\left(\frac{\pi}{6}\right) & 0 \\
0 & 0 & 1
\end{bmatrix}
\end{equation}

\begin{equation}
T_2 = \begin{bmatrix}
1 & 0 & 0 \\
0 & \cos\left(\frac{\pi}{6}\right) & \sin\left(\frac{\pi}{6}\right) \\
0 & -\sin\left(\frac{\pi}{6}\right) & \cos\left(\frac{\pi}{6}\right)
\end{bmatrix}
\end{equation}

\begin{equation}
T_3 = \begin{bmatrix}
\cos\left(\frac{\pi}{6}\right) & 0 & -\sin\left(\frac{\pi}{6}\right) \\
0 & 1 & 0 \\
\sin\left(\frac{\pi}{6}\right) & 0 & \cos\left(\frac{\pi}{6}\right)
\end{bmatrix}
\end{equation}
For each dataset, we experimented with each rotation augmentation separately and a combination of all. The rotation with the best results was selected. 

\noindent To introduce bias-based augmentation, bias values were randomly sampled from \( b \sim \mathcal{N}(0, \textit{bias\_std}) \), simulating systematic offsets. The standard deviation (\textit{bias\_std}) was set differently for the accelerometer and gyroscope: \( 0.1\ \mathrm{m/s^2} \) for accelerometer data and \( 0.001\ \mathrm{rad/s} \) for gyroscope data. Two experimental setups were evaluated: in the first, a single bias value was sampled, while in the second, three bias values were sampled from the same distribution. The better results from these experiments were selected for further analysis.

\noindent Similarly, for noise-based augmentation, noise values were randomly sampled from \( n \sim \mathcal{N}(0, \textit{noise\_std}) \), mimicking random fluctuations. The standard deviation (\textit{noise\_std}) was also differentiated for the accelerometer and gyroscope, set to \( 0.1\ \mathrm{m/s^2} \) and \( 0.001\ \mathrm{rad/s} \), respectively. Two experimental setups were examined: in the first, a single noise value was sampled, and in the second, three noise values were sampled with incrementally increasing standard deviations, scaling up by one-quarter increments from the initial value. For example, the accelerometer standard deviations were \( 0.1 \), \( 0.25 \), and \( 0.5 \). The best-performing results from these experiments were selected for analysis.
A comprehensive summary of all rotation, bias, and noise parameters is provided in Table~\ref{tbl:aug_params}.

\begin{figure*}[!htb]
    \centering
    \includegraphics[width=\textwidth]{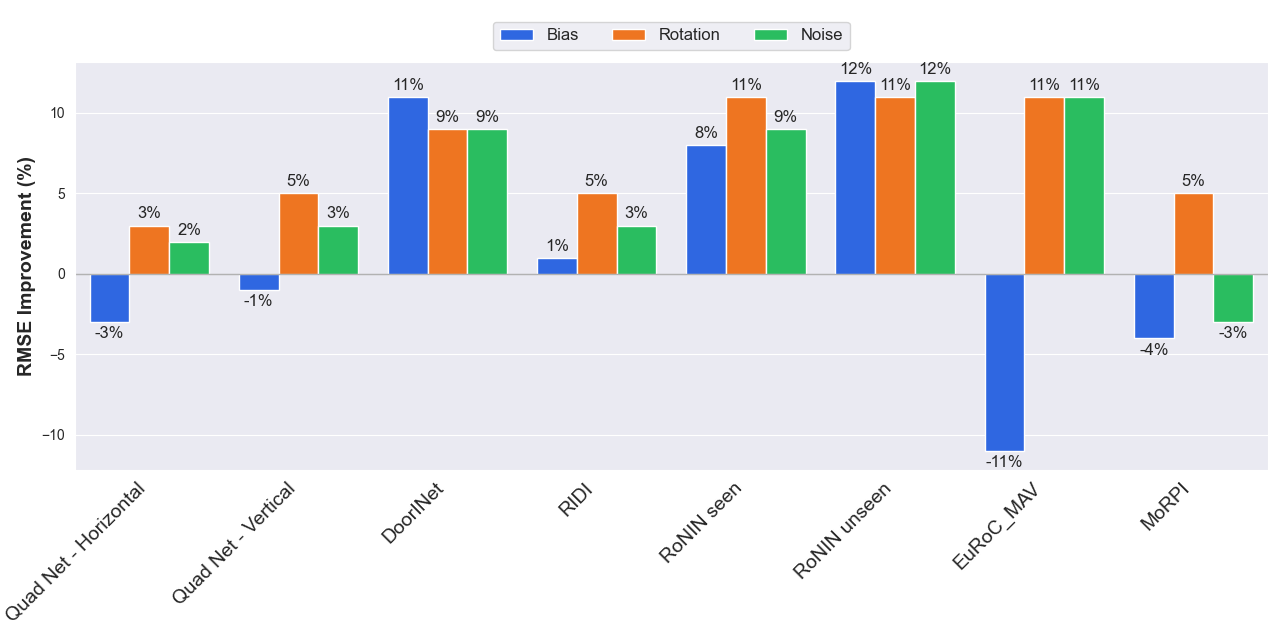}
    \caption{\small RMSE improvement in percentages for each dataset using data augmentation approaches.}
    \label{fig:aug_res_sum}
\end{figure*}

\noindent The results across all datasets are presented in Figure \ref{fig:aug_res_sum}. 
Rotation augmentation resulted in improvements across all datasets, achieving an average gain of 7\%, bias augmentation showed gains in four out of eight datasets, with an average improvement of 2\%, and noise augmentation enhanced performance in seven out of eight datasets, averaging a 6\% increase. 
Although bias augmentation performed worse on average compared to rotation and noise augmentation, it outperformed both in specific cases, such as the DoorINet dataset with an 11\% improvement and the RoNIN unseen dataset with a 12\% improvement. 

\begin{table}[!htbp]
\caption{Rotation, bias and noise augmentation parameters for each dataset.}
\begin{center}
\begin{adjustbox} {width = 1\columnwidth}
\begin{tabular}{|c|c|c|c|}
\hline
\makecell{\textbf{Dataset}} & \makecell{\textbf{Rotation}} & \makecell{\textbf{Bias Experiment}} & \makecell{\textbf{Noise Experiment}}
\\ \hline
QuadNet - Horizontal & $T_1$ & $b \times 1$ & $n \times 1$        \\
QuadNet - Vertical & $T_1$ & $b \times 1$ & $n \times 1$      \\
EuRoC\_MAV & $T_2$ & $b \times 1$     & $n \times 1$     \\
DoorINet   & $T_2$ & $b \times 3$      & $n \times 3$        \\
RIDI       & $T_3$ & $b \times 3$       & $n \times 3$     \\
RoNIN - seen  & $T_3$ & $b \times 3$     &  $n \times 3$    \\
RoNIN - unseen & $T_3$ & $b \times 3$    &  $n \times 3$    \\
MoRPI      & $T_2$ & $b \times 1$     &  $n \times 1$    \\
\hline
\end{tabular}
\label{tbl:aug_params}
\end{adjustbox}
\end{center}
\vspace{0.1em} 
\noindent \textit{Note:} $T_1$, $T_2$, and $T_3$ represent different rotation matrices. The terms $b \times 1$, $b \times 3$, $n \times 1$, and $n \times 3$  indicate the number of biases and noises sampled in the respective experiments.
\end{table}

\subsection{Data Preprocessing}
\subsubsection{Inertial Noise} \label {res:inertial_noise}
For the denoising method, three different window sizes were evaluated for each dataset. The window size yielding the best results was then selected. The denoising window size parameters are detailed in Table \ref{tbl:noise_parameters}.
In the case of noise addition, a consistent mean of zero, a standard deviation of $0.1\  \mathrm{m/s^2}$ for accelerometer data, and a standard deviation of $0.001\ \mathrm{rad/s}$ were maintained in all datasets.

\begin{figure}[!htb]
    \centering
    \includegraphics[width=\linewidth]{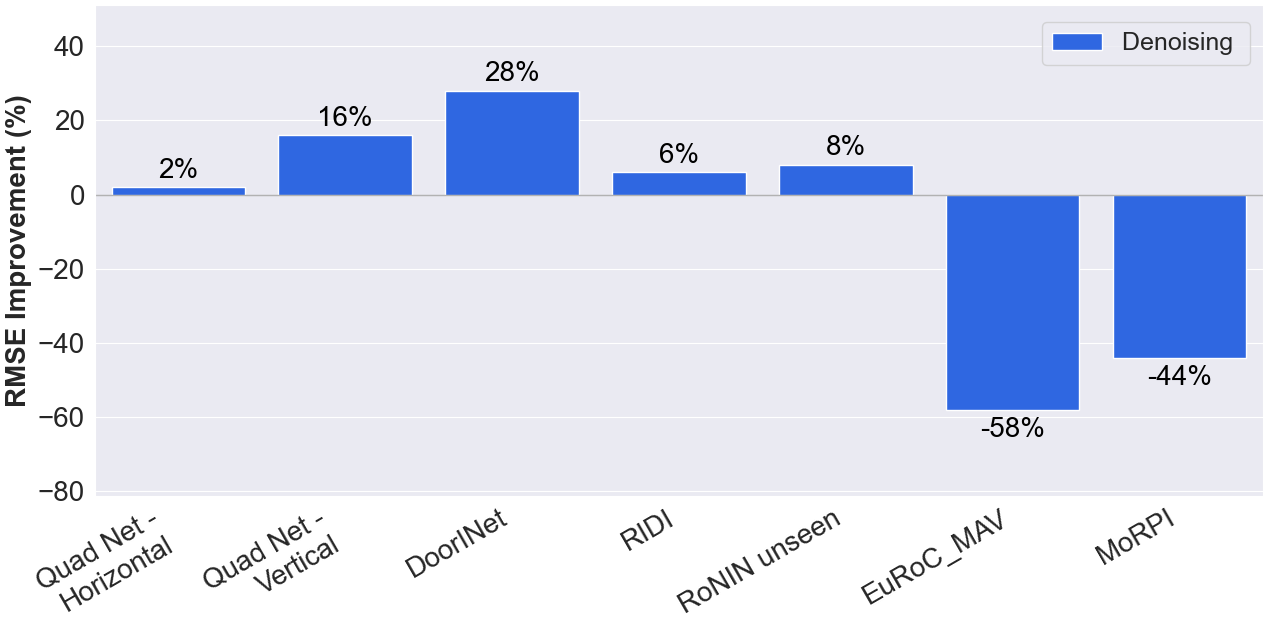}
    \caption{\small RMSE improvement in percentages for each dataset using data noise processing techniques.}
    \label{fig:noise_res_sum}
\end{figure}

\noindent A summary of the denoising RMSE improvement results is provided in Figure \ref{fig:noise_res_sum}. 
Adding noise resulted in small improvements in five out of eight datasets, with the largest gain being 3\% on the RoNIN datasets and an overall average improvement of 1\%, and therefore not presented in the figure. Denoising, showed improvements in five out of eight datasets and had an average decrease of 5\%.
Yet, certain datasets exhibited significant gains, including a 28\% improvement on DoorINet, 16\% on QuadNet - Vertical, and 8\% on RoNIN unseen. However, it led to substantial performance declines on the EuRoC\_MAV and MoRPI datasets, with reductions of -58\% and -44\%, respectively.

\begin{table}[htbp]
\caption{Denoising window size values across all datasets.}
\begin{center}
\begin{tabular}{|c|c|}
\hline
\textbf{Dataset} & \textbf{Window Size} \\
\hline
QuadNet - Horizontal    & 50  \\
QuadNet - Vertical    & 50  \\
EuRoC\_MAV & 10  \\
DoorINet   & 50  \\
RIDI       & 50  \\
RoNIN - seen      & 25  \\
RoNIN - unseen     & 25  \\
MoRPI      & 10  \\

\hline
\end{tabular}
\label{tbl:noise_parameters}
\end{center}
\end{table}

\subsubsection{Data Normalization} \label {res:norm}
Two normalization methods, standard normalization and robust normalization, were evaluated, both of which demonstrated generally poor performance. The results of the normalization techniques are summarized in Table \ref{tbl:norm_results}.
Standart normalization showed improvements in two out of eight datasets, with an overall average change of -308\% and robust normalization, showed improvements in one out of eight datasets and had an average decrease of -907\%.
Notably, there are exceptional cases, such as in the RoNIN dataset, where standard and robust normalization enhances performance by up to 11\%. This can be attributed to the fact that standard normalization retains the signal's overall pattern and magnitude while effectively filtering out unwanted noise.

\begin{table}[htbp]
\caption{Data normalization results showing RMSE improvement in percentages.}
\begin{center}
\begin{tabular}{|c|c|c|c|}
\hline
\textbf{Dataset} & \textbf{Standart} & \textbf{Robust} \\
\hline
QuadNet - Horizontal    & -23\%  & -34\%  \\
QuadNet - Vertical     & -11\%  & -26\%  \\
DoorINet     & -2222\% & -7001\%  \\
RIDI        & -2\%  & -5\%  \\
RoNIN seen       & 2\%  & -4\% \\
RoNIN unseen     & 11\%  & 3\% \\
EuRoC\_MAV   & -193\%  & -130\% \\
MoRPI   & -25\% & -55\% \\ 
\hline
\end{tabular}
\label{tbl:norm_results}
\end{center}
\end{table}

\subsubsection{Detrending} \label {res:detrend}
\begin{figure}[!htb]
    \centering
    \includegraphics[width=\linewidth]{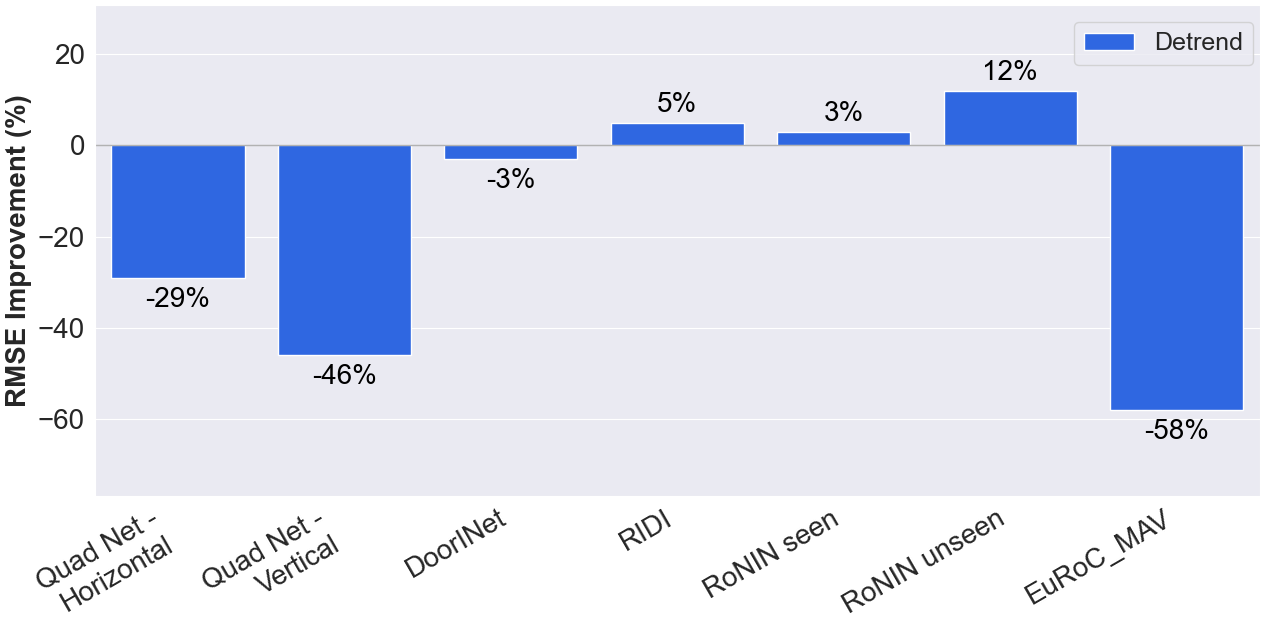}
    \caption{\small RMSE improvement in percentages for each dataset using data linear detrending techniques.}
    \label{fig:detrend_res}
\end{figure}

The linear detrending results are summarized in Figure \ref{fig:detrend_res}. 
Linear detrending improved performance in three out of eight datasets, with an overall average change of -17\%. Despite the significant average decline, notable gains were observed in pedestrian datasets such as RoNIN and RIDI, including a 12\% improvement for the RoNIN unseen subset, a 3\% improvement for the RoNIN seen subset, and a 5\% improvement for the RIDI dataset.

\begin{figure*}[!htbp]
    \centering
    \includegraphics[width=15 cm]{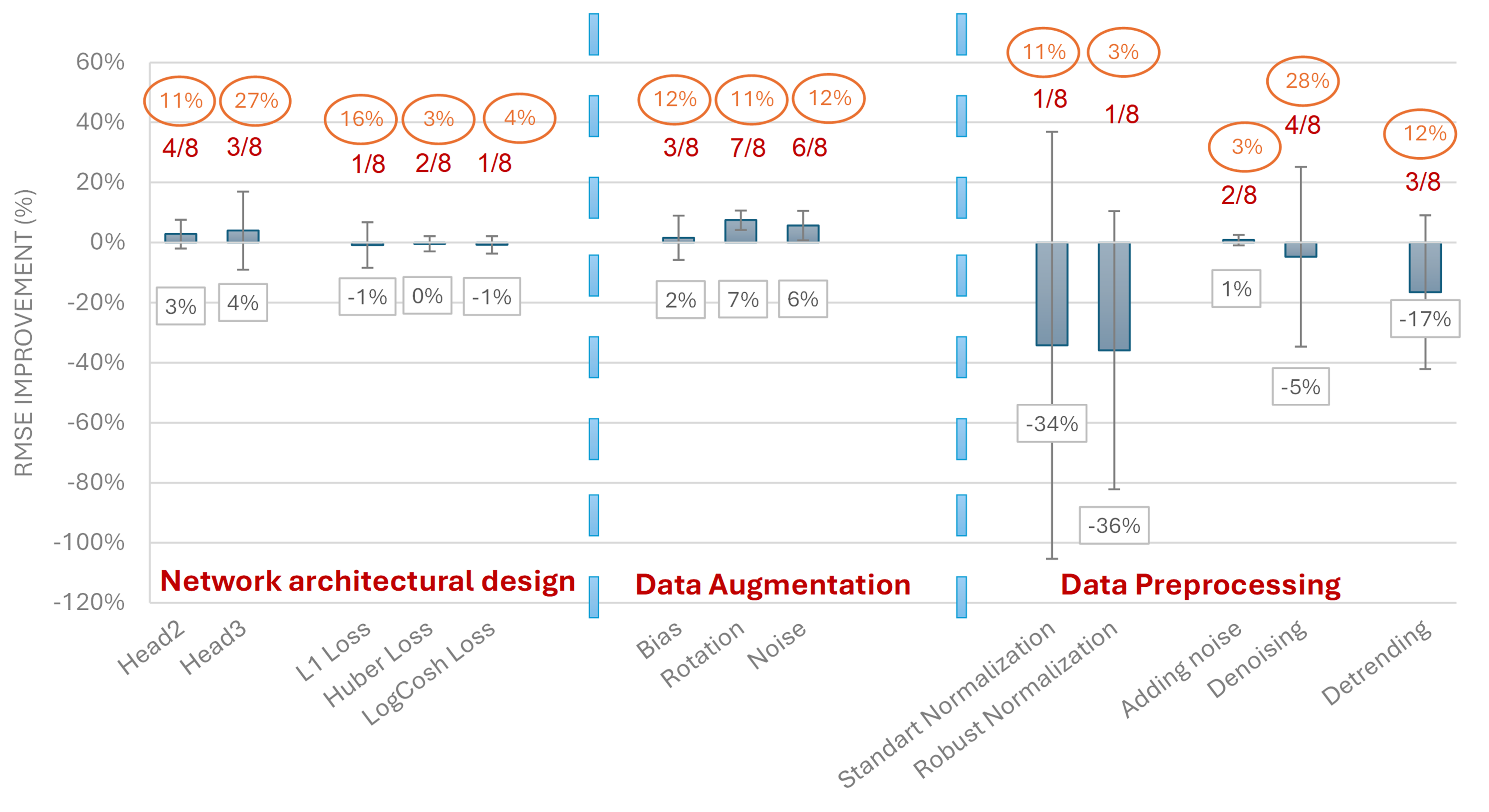}
    \caption{\small Average and standard deviation obtained for all datasets in each experiment.  The numbers above the bars indicate in how many datasets the RMSE improvement was over 3\%. The highest improvement value is presented inside a circle. The normalization average excludes DoorINet results due to their exceptionally large percentages.}
    \label{fig:sum_all}
\end{figure*}
\subsection{Summary} \label{res:summary} 
\noindent A summary of the main results across all eight datasets and 13 experiments is presented in Figure \ref{fig:sum_all}. It presents the average and standard deviation obtained for all datasets in each experiment. The numbers above the bars represent the count of datasets where the RMSE improvement exceeded 3\%. The highest improvement value is highlighted within a circle.
The most effective techniques for improving accuracy were data augmentation through rotation and noise addition, achieving average improvements of 7\% and 6\%, respectively, with consistent gains across most datasets.
The denoising results highlight a key finding: while the overall average showed a 5\% decline, four datasets experienced significant improvements of 28\%, 16\%, 8\%, and 6\% for DoorINet, QuadNet-Vertical, RoNIN-unseen, and RIDI datasets, respectively.
The multi-head models also demonstrated notable performance enhancements. The Head2 model achieved an average improvement of 3\%, showing greater robustness with significant gains across four datasets and no major declines. The Head3 model exhibited a smaller average improvement of 2\% and with significant performance reductions in three datasets, indicating less stability.
Normalization techniques consistently led to substantial performance reductions, with average decreases of 308\% and 907\% observed across the two methods, highlighting the fact that the actual inertial measurement should not be normalized.
A comparison of different loss functions revealed minimal overall impact, with average changes of -1\%, 0\%, and -1\% for the L1, Huber, and LogCosh loss functions compared to MSE. However, as noted in the relevant section, specific cases may favor one loss function over others.
The noise addition technique had no significant overall effect, with a modest 1\% improvement, though notable gains were observed in the RoNIN dataset for both seen and unseen cases.
Finally, detrending yielded mixed outcomes. While the overall average performance decreased by 17\%, pedestrian datasets experienced significant improvements of 12\%, 3\%, and 5\% in the RoNIN-unseen, RoNIN-seen, and RIDI datasets, respectively.

\section{Conclusions} \label{sec:conclusions}
\noindent This study presented a comprehensive evaluation of 13 techniques for improving the accuracy of deep learning models applied to inertial data problems. Three different aspects were explored: network architectural design, data augmentation and data preprocessing across eight distinct datasets: QuadNet - Horizontal, QuadNet - Vertical, EuRoC\_MAV, DoorINet, RIDI, RoNIN - seen, RoNIN - unseen, and MoRPI. Those datasets include 1079 minutes of inertial data sampled between 120-200Hz.
Also, datasets include different platforms, such as quadrotors, doors, pedestrians, and mobile robots, used for a variety of inertial applications.
Our findings demonstrate that data augmentation methods, particularly rotation and noise addition, emerged as the most effective in improving performance across datasets, achieving average gains of 7\% and 6\%, respectively. These techniques introduced variability and resilience to the models without requiring extensive new data collection.
Multi-head architectures, Head2 and Head3 models demonstrated clear benefits over single-head networks in processing inertial data. The Head2 model provided more consistent results across datasets, reflecting its ability to balance accuracy with stability. 
The selection of loss functions had minimal overall impact, though specific cases demonstrated the utility of alternative loss functions like Huber and L1.
Data preprocessing methods such as detrending and denoising yielded mixed results, emphasizing their dependence on specific dataset characteristics and application contexts. Detrending showed notable improvements for pedestrian datasets, while denoising produced significant gains for certain datasets but led to large performance declines in others. Additive noise, while was recommended to enhance model robustness, showed only marginal benefits in this study, indicating that its utility may be limited. 
Normalization techniques led to significant performance degradation across most datasets, reflecting the complexity and sensitivity of inertial data to such transformations. \\
In conclusion, this study presents effective strategies to improve the accuracy of inertial-based systems through deep learning and underscores the importance of dataset-specific tuning. Additionally, this study
outlines benchmarking strategies for enhancing neural inertial regression networks.\\
Future research should aim to refine these techniques and explore other techniques to improve the accuracy and reliability of inertial sensing systems in diverse real-world applications. Additionally, testing a broader range of datasets and applying these methods to classification tasks could further validate the presented conclusions and uncover new insights.

\section*{Acknowledgment}
N. P-H., N. P., and A. E. are supported by the Maurice Hatter Foundation.

\bibliographystyle{IEEEtran}
\bibliography{./references.bib}

\begin{thebibliography}{10}
\providecommand{\url}[1]{#1}
\csname url@samestyle\endcsname
\providecommand{\newblock}{\relax}
\providecommand{\bibinfo}[2]{#2}
\providecommand{\BIBentrySTDinterwordspacing}{\spaceskip=0pt\relax}
\providecommand{\BIBentryALTinterwordstretchfactor}{4}
\providecommand{\BIBentryALTinterwordspacing}{\spaceskip=\fontdimen2\font plus
\BIBentryALTinterwordstretchfactor\fontdimen3\font minus \fontdimen4\font\relax}
\providecommand{\BIBforeignlanguage}[2]{{%
\expandafter\ifx\csname l@#1\endcsname\relax
\typeout{** WARNING: IEEEtran.bst: No hyphenation pattern has been}%
\typeout{** loaded for the language `#1'. Using the pattern for}%
\typeout{** the default language instead.}%
\else
\language=\csname l@#1\endcsname
\fi
#2}}
\providecommand{\BIBdecl}{\relax}
\BIBdecl

\bibitem{groves2015navigation}
P.~D. Groves, ``{Principles of {GNSS}, inertial, and multisensor integrated navigation systems, [Book review]},'' \emph{IEEE Aerospace and Electronic Systems Magazine}, vol.~30, no.~2, pp. 26--27, 2015.

\bibitem{cuesta2010use}
A.~I. Cuesta-Vargas, A.~Gal{\'a}n-Mercant, and J.~M. Williams, ``{The use of inertial sensors system for human motion analysis},'' \emph{Physical Therapy Reviews}, vol.~15, no.~6, pp. 462--473, 2010.

\bibitem{engelsman2022measurement}
D.~Engelsman, T.~Sherif, S.~Meller, F.~Twele, I.~Klein, A.~Zamansky, and H.~A. Volk, ``{Measurement of canine ataxic gait patterns using body-worn smartphone sensor data},'' \emph{Frontiers in veterinary science}, vol.~9, p. 912253, 2022.

\bibitem{cohen2024inertial}
N.~Cohen and I.~Klein, ``{Inertial navigation meets deep learning: A survey of current trends and future directions},'' \emph{Results in Engineering}, p. 103565, 2024.

\bibitem{klein2022data}
I.~Klein, ``{Data-driven meets navigation: Concepts, models, and experimental validation},'' in \emph{2022 DGON Inertial Sensors and Systems (ISS)}.\hskip 1em plus 0.5em minus 0.4em\relax IEEE, 2022, pp. 1--21.

\bibitem{camomilla2018trends}
V.~Camomilla, E.~Bergamini, S.~Fantozzi, and G.~Vannozzi, ``{Trends supporting the in-field use of wearable inertial sensors for sport performance evaluation: A systematic review},'' \emph{Sensors}, vol.~18, no.~3, p. 873, 2018.

\bibitem{worsey2019inertial}
M.~T. Worsey, H.~G. Espinosa, J.~B. Shepherd, and D.~V. Thiel, ``{Inertial sensors for performance analysis in combat sports: A systematic review},'' \emph{Sports}, vol.~7, no.~1, p.~28, 2019.

\bibitem{shaeffer2013mems}
D.~K. Shaeffer, ``{{MEMS} inertial sensors: A tutorial overview},'' \emph{IEEE Communications Magazine}, vol.~51, no.~4, pp. 100--109, 2013.

\bibitem{sarker2021deep}
I.~H. Sarker, ``{Deep learning: a comprehensive overview on techniques, taxonomy, applications and research directions},'' \emph{SN computer science}, vol.~2, no.~6, p. 420, 2021.

\bibitem{malhotra2023recent}
R.~Malhotra and P.~Singh, ``{Recent advances in deep learning models: a systematic literature review},'' \emph{Multimedia Tools and Applications}, vol.~82, no.~29, pp. 44\,977--45\,060, 2023.

\bibitem{chen2024deep}
C.~Chen and X.~Pan, ``{Deep learning for inertial positioning: A survey},'' \emph{IEEE Transactions on Intelligent Transportation Systems}, 2024.

\bibitem{li2022calib}
R.~Li, C.~Fu, W.~Yi, and X.~Yi, ``{Calib-Net: Calibrating the low-cost IMU via deep convolutional neural network},'' \emph{Frontiers in Robotics and AI}, vol.~8, p. 772583, 2022.

\bibitem{engelsman2023data}
D.~Engelsman and I.~Klein, ``{Data-driven denoising of stationary accelerometer signals},'' \emph{Measurement}, vol. 218, p. 113218, 2023.

\bibitem{asraf2021pdrnet}
O.~Asraf, F.~Shama, and I.~Klein, ``{{PDRNet}: A deep-learning pedestrian dead reckoning framework},'' \emph{IEEE Sensors Journal}, vol.~22, no.~6, pp. 4932--4939, 2021.

\bibitem{chen2020deep}
C.~Chen, P.~Zhao, C.~X. Lu, W.~Wang, A.~Markham, and N.~Trigoni, ``{Deep-learning-based pedestrian inertial navigation: Methods, data set, and on-device inference},'' \emph{IEEE Internet of Things Journal}, vol.~7, no.~5, pp. 4431--4441, 2020.

\bibitem{zhang2020navnet}
X.~Zhang, B.~He, G.~Li, X.~Mu, Y.~Zhou, and T.~Mang, ``{NavNet}: {AUV} navigation through deep sequential learning,'' \emph{IEEE Access}, vol.~8, pp. 59\,845--59\,861, 2020.

\bibitem{or2023pronet}
B.~Or and I.~Klein, ``{{ProNet}: Adaptive process noise estimation for {INS/DVL} fusion},'' in \emph{2023 IEEE Underwater Technology (UT)}.\hskip 1em plus 0.5em minus 0.4em\relax IEEE, 2023, pp. 1--5.

\bibitem{solodar2024vio}
D.~Solodar and I.~Klein, ``{{VIO}-{DualProNet}: Visual-inertial odometry with learning based process noise covariance},'' \emph{Engineering Applications of Artificial Intelligence}, vol. 133, p. 108466, 2024.

\bibitem{esfahani2019aboldeepio}
M.~A. Esfahani, H.~Wang, K.~Wu, and S.~Yuan, ``{{AbolDeepIO}: A novel deep inertial odometry network for autonomous vehicles},'' \emph{IEEE Transactions on Intelligent Transportation Systems}, vol.~21, no.~5, pp. 1941--1950, 2019.

\bibitem{bengio2013deep}
Y.~Bengio, ``{Deep learning of representations: Looking forward},'' in \emph{International conference on statistical language and speech processing}.\hskip 1em plus 0.5em minus 0.4em\relax Springer, 2013, pp. 1--37.

\bibitem{kawaguchi2017generalization}
K.~Kawaguchi, L.~P. Kaelbling, and Y.~Bengio, ``{Generalization in deep learning},'' \emph{arXiv preprint arXiv:1710.05468}, vol.~1, no.~8, 2017.

\bibitem{silva2019end}
J.~P. Silva~do Monte~Lima, H.~Uchiyama, and R.-i. Taniguchi, ``{End-to-end learning framework for {IMU}-based 6-dof odometry},'' \emph{Sensors}, vol.~19, no.~17, p. 3777, 2019.

\bibitem{kingma2014adam}
D.~P. Kingma, ``{Adam: A method for stochastic optimization},'' \emph{arXiv preprint arXiv:1412.6980}, 2014.

\bibitem{herath2020ronin}
S.~Herath, H.~Yan, and Y.~Furukawa, ``{{RoNIN}: Robust neural inertial navigation in the wild: Benchmark, evaluations, \& new methods},'' in \emph{2020 IEEE international conference on robotics and automation (ICRA)}.\hskip 1em plus 0.5em minus 0.4em\relax IEEE, 2020, pp. 3146--3152.

\bibitem{shavit2021boosting}
Y.~Shavit and I.~Klein, ``{Boosting inertial-based human activity recognition with transformers},'' \emph{IEEE Access}, vol.~9, pp. 53\,540--53\,547, 2021.

\bibitem{liu2023smartphone}
F.~Liu, H.~Ge, D.~Tao, R.~Gao, and Z.~Zhang, ``{Smartphone-based Pedestrian Inertial Tracking: Dataset, Model, and Deployment},'' \emph{IEEE Transactions on Instrumentation and Measurement}, 2023.

\bibitem{wang2020comprehensive}
Q.~Wang, Y.~Ma, K.~Zhao, and Y.~Tian, ``{A comprehensive survey of loss functions in machine learning},'' \emph{Annals of Data Science}, pp. 1--26, 2020.

\bibitem{he2016deep}
K.~He, X.~Zhang, S.~Ren, and J.~Sun, ``{Deep residual learning for image recognition},'' in \emph{Proceedings of the IEEE conference on computer vision and pattern recognition}, 2016, pp. 770--778.

\bibitem{um2017data}
T.~T. Um, F.~M. Pfister, D.~Pichler, S.~Endo, M.~Lang, S.~Hirche, U.~Fietzek, and D.~Kuli{\'c}, ``{Data augmentation of wearable sensor data for parkinson’s disease monitoring using convolutional neural networks},'' in \emph{Proceedings of the 19th ACM international conference on multimodal interaction}, 2017, pp. 216--220.

\bibitem{khaddour2021survey}
M.~Khaddour, S.~Shidlovskiy, D.~Shashev, and M.~Mondal, ``{Survey of Denoising Methods for Inertial Sensor Measurements},'' in \emph{2021 International Conference on Information Technology (ICIT)}.\hskip 1em plus 0.5em minus 0.4em\relax IEEE, 2021, pp. 787--790.

\bibitem{gonzalez2018statistical}
R.~Gonzalez and C.~A. Catania, ``{A statistical approach for optimal order adjustment of a moving average filter},'' in \emph{2018 IEEE/ION Position, Location and Navigation Symposium (PLANS)}.\hskip 1em plus 0.5em minus 0.4em\relax IEEE, 2018, pp. 1542--1546.

\bibitem{esfahani2019orinet}
M.~A. Esfahani, H.~Wang, K.~Wu, and S.~Yuan, ``{{OriNet}: Robust {3-D} orientation estimation with a single particular {IMU}},'' \emph{IEEE Robotics and Automation Letters}, vol.~5, no.~2, pp. 399--406, 2019.

\bibitem{cabello2023impact}
K.~Cabello-Solorzano, I.~Ortigosa~de Araujo, M.~Pe{\~n}a, L.~Correia, and A.~J.~Tall{\'o}n-Ballesteros, ``{The impact of data normalization on the accuracy of machine learning algorithms: a comparative analysis},'' in \emph{International Conference on Soft Computing Models in Industrial and Environmental Applications}.\hskip 1em plus 0.5em minus 0.4em\relax Springer, 2023, pp. 344--353.

\bibitem{lima2023large}
F.~T. Lima and V.~M. Souza, ``{A large comparison of normalization methods on time series},'' \emph{Big Data Research}, vol.~34, p. 100407, 2023.

\bibitem{hurwitz2023quadrotor}
D.~Hurwitz and I.~Klein, ``{Quadrotor Dead Reckoning with Multiple Inertial Sensors},'' in \emph{2023 DGON Inertial Sensors and Systems (ISS)}.\hskip 1em plus 0.5em minus 0.4em\relax IEEE, 2023, pp. 1--18.

\bibitem{hurwitz2024deep}
D.~Hurwitz, N.~Cohen, and I.~Klein, ``Deep-learning-assisted inertial dead reckoning and fusion,'' \emph{IEEE Transactions on Instrumentation and Measurement}, vol.~74, pp. 1--9, 2025.

\bibitem{burri2016euroc}
M.~Burri, J.~Nikolic, P.~Gohl, T.~Schneider, J.~Rehder, S.~Omari, M.~W. Achtelik, and R.~Siegwart, ``{The {EuRoC} micro aerial vehicle datasets},'' \emph{The International Journal of Robotics Research}, vol.~35, no.~10, pp. 1157--1163, 2016.

\bibitem{achtelik2012design}
M.~Achtelik, K.-M. Doth, D.~Gurdan, and J.~Stumpf, ``{Design of a multi rotor {MAV} with regard to efficiency, dynamics and redundancy},'' in \emph{AIAA Guidance, Navigation, and Control Conference}, 2012, p. 4779.

\bibitem{zakharchenko2024doorinet}
A.~Zakharchenko, S.~Farber, and I.~Klein, ``{{DoorINet}: Door Heading Prediction through Inertial Deep Learning},'' \emph{IEEE Sensors Journal}, 2024.

\bibitem{yan2018ridi}
H.~Yan, Q.~Shan, and Y.~Furukawa, ``{{RIDI}: Robust {IMU} double integration},'' in \emph{Proceedings of the European conference on computer vision (ECCV)}, 2018, pp. 621--636.

\bibitem{etzion2024snake}
A.~Etzion and I.~Klein, ``{Snake-Inspired Mobile Robot Positioning with Hybrid Learning},'' \emph{arXiv preprint arXiv:2411.17430}, 2024.

\bibitem{zhuang2022randomness}
D.~Zhuang, X.~Zhang, S.~Song, and S.~Hooker, ``{Randomness in neural network training: Characterizing the impact of tooling},'' \emph{Proceedings of Machine Learning and Systems}, vol.~4, pp. 316--336, 2022.

\end{thebibliography}

\end{document}